\documentclass[10pt,twocolumn]{article}
\usepackage[a4paper,margin=1in]{geometry}
\usepackage[sort&compress,numbers]{natbib}
\usepackage{graphicx}
\usepackage{amsmath,amssymb}
\usepackage{booktabs}
\usepackage{multirow}
\usepackage{colortbl}
\usepackage{xcolor}
\usepackage{tabularx}
\usepackage{array}
\usepackage[table]{xcolor}
\usepackage{multirow}
\usepackage{makecell}   
\usepackage{colortbl}  
\usepackage{subcaption}
\usepackage{pifont}
\usepackage{lineno}
\usepackage[hidelinks]{hyperref}
\usepackage{titlesec}
\titlespacing*{\section}{0pt}{0.7ex plus 0.2ex minus 0.1ex}{0.5ex plus 0.1ex}
\titlespacing*{\subsection}{0pt}{0.55ex plus 0.15ex minus 0.08ex}{0.4ex plus 0.08ex}
\titlespacing*{\subsubsection}{0pt}{0.45ex plus 0.12ex minus 0.06ex}{0.3ex plus 0.06ex}
\definecolor{lightgrayborder}{RGB}{200, 200, 200}
\newcommand{\cmark}{\ding{51}}
\newcommand{\xmark}{\ding{55}}
\title{AU-Guided Synthetic Video Generation for Micro-Expression Recognition}
\author{
Pei-Sze Tan \and
Sailaja Rajanala \and
Yee-Fan Tan \and
Rapha\"{e}l C.-W. Phan \and
Huey-Fang Ong \\
CyPhi AI Lab, Monash University Malaysia
}
\date{}

\begin{document}

\def\floatpagepagefraction{1}
\def\textpagefraction{.001}

\maketitle

\begin{abstract}
Micro-expression recognition is limited by the small scale, narrow demographic coverage, and restricted emotion labels of existing datasets. We introduce \textbf{EquiME}, a synthetic micro-expression dataset built from AU-guided image-to-video generation. EquiME contains 75K videos generated from 15K source face images across five target emotions, together with automatically inferred demographic metadata and video-quality measurements. We evaluate EquiME using frame-pair similarity, spatial variation, and no-reference perceptual-quality metrics, together with cross-dataset MER experiments on SAMM and CASME II. Models trained on EquiME achieve competitive cross-dataset performance on SAMM and CASME II and show comparatively low variation across the four evaluated architectures. This paper focuses on the dataset design, the structured AU-conditioning pipeline used for video generation, and the empirical evidence needed to assess EquiME as a synthetic MER resource. Project page: \href{https://kirito-blade.github.io/me-vlm/}{https://kirito-blade.github.io/me-vlm/}
\end{abstract}

\begin{center}
\end{center}
\begin{figure*}[t]
  \centering
  \includegraphics[width=\textwidth]{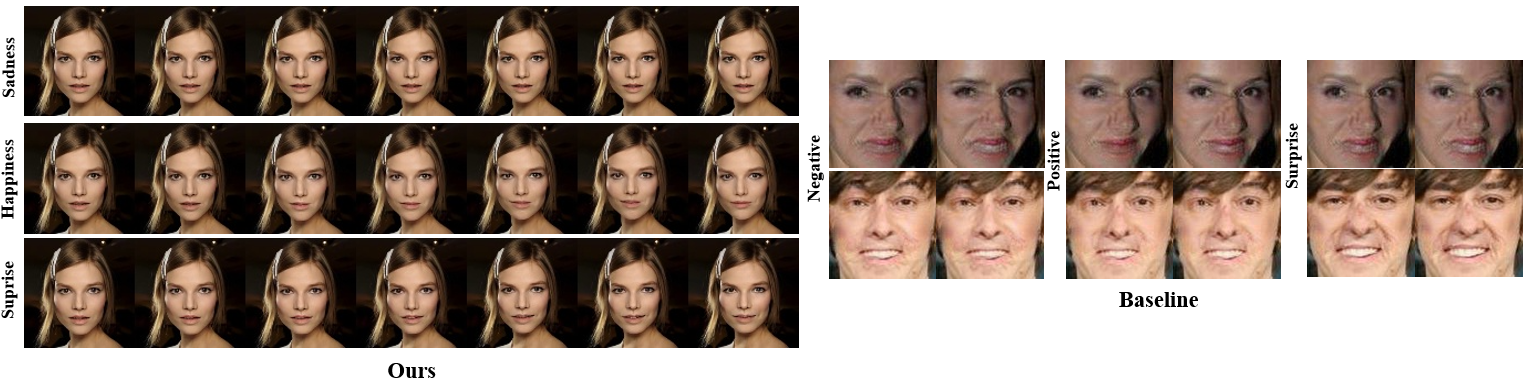}
  \caption{EquiME provides temporally coherent video clips rather than isolated key frames and includes multiple prompted expressions generated from the same source face.}
  \label{fig:teaser}
\end{figure*}
\section{Introduction}
Micro-expression recognition (MER) studies subtle and short-lived facial movements that can reveal underlying affective states. The problem is important for affective computing, human-computer interaction, and behavioral analysis, but progress remains constrained by the datasets available for training and evaluation \cite{surveydeep, xie2022overview}. Compared with mainstream facial analysis tasks, MER still relies on small corpora collected in controlled environments, often with limited subject diversity and uneven emotion coverage \cite{casmeii,casmesquare,casmecube,samm,sammlv,smic,smice}. These limitations make it difficult to train robust models and to assess whether performance generalizes across identities and demographic groups.

The data bottleneck is not only about scale. Existing benchmark datasets are also narrow in demographic coverage, which can increase the risk that MER systems overfit to dataset-specific appearance cues rather than learn expression dynamics that transfer across populations \cite{dominguez2024metrics, xu2020investigating}. This concern is especially relevant in MER because the signal of interest is weak: subtle muscle activations can easily be confounded by identity, lighting, recording conditions, and annotation noise. Standard evaluation settings such as leave-one-subject-out validation partly address subject overlap, but they do not fully resolve broader questions of generalization across demographic groups \cite{austin2025distributional, gronau2019limitations}.

Synthetic data offers a practical way to expand MER training resources, but current synthetic alternatives remain limited. Prior work has explored image-based synthesis or AU-driven generation \cite{yap2021synthesising, zhang2023facial, zhao2022fine, miex}, yet available datasets often trade off realism, controllability, temporal continuity, or demographic coverage. In particular, existing open synthetic resources do not provide the combination of temporally coherent micro-expression videos, source-face diversity, and rich metadata needed for rigorous cross-dataset evaluation.

\paragraph{\textbf{Overview of EquiME.}} We present \textbf{EquiME}, a synthetic micro-expression video dataset designed for cross-dataset MER research. EquiME is generated from source face images using an AU-guided image-to-video pipeline. The key idea is to treat facial Action Units (AUs) as an intermediate control representation: they provide a compact way to specify subtle muscle activations while preserving an explicit mapping between facial movement descriptions and target emotion labels. Using this formulation, we generate short video clips that cover the temporal progression of an expression while maintaining source-face consistency across samples.

\begin{figure*}[t]
    \centering
    \includegraphics[width=\textwidth]{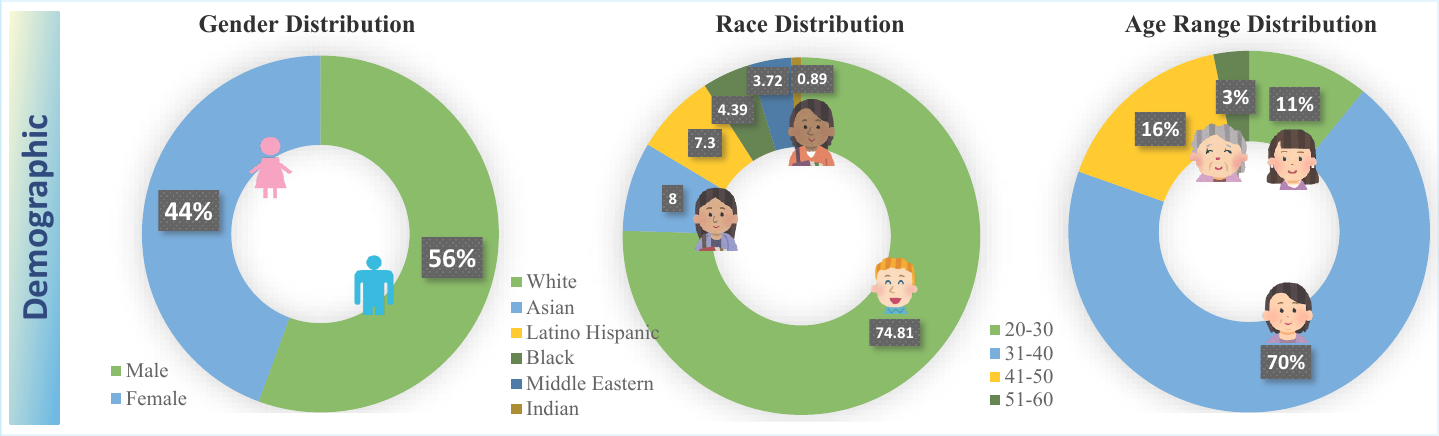}
    \caption{Demographic summary of EquiME. The release includes automatically inferred Male/Female labels, estimated demographic-category labels, and estimated age-group labels.}
    \label{fig:mevlm_dist}
\end{figure*}

EquiME contains 75K synthetic videos generated from 15K source face images spanning five emotion classes: happiness, sadness, surprise, disgust, and anger. The dataset is accompanied by automatically inferred demographic metadata and video-quality measurements that may support controlled subgroup analysis and generalization studies. Figure~\ref{fig:mevlm_dist} summarizes the current demographic distribution of the release. The dataset is not perfectly balanced across all groups, but it provides broader coverage than several commonly used MER benchmarks and supports controlled subset construction for future studies.

Table~\ref{tab:dataset_comparison} positions EquiME against representative real-world MER datasets and MiE-X \cite{miex}, an open synthetic alternative. Unlike small real-world benchmarks, EquiME scales to substantially more source face images and generated samples. Unlike frame-based synthetic datasets, it provides full video sequences rather than isolated key frames, which is useful for studying temporal facial-motion patterns in downstream MER models.

Our contributions are:
\begin{enumerate}
\item We introduce EquiME, a large synthetic MER video dataset with 75K clips generated from 15K source face images, five emotion categories, and associated demographic metadata.
\item We develop an AU-guided image-to-video generation pipeline that uses structured prompts and negative constraints to produce temporally coherent, micro-expression-oriented facial-motion clips.
\item We formalize the generation process with a structured AU-conditioning formulation that makes the prompt-construction and video-generation stages explicit.
\item We evaluate EquiME using both perceptual video-quality metrics and cross-dataset MER experiments, and we report that it transfers reasonably well to the real MER benchmarks considered in this paper.
\end{enumerate}

\begin{table*}[t]
\centering
\caption{Comparison with representative real-human ME datasets$^\dagger$ (CASME II~\cite{casmeii}, SAMM~\cite{samm}, SMIC~\cite{smic}) and synthetic MiE-X~\cite{miex}. $^\ddagger$Generated temporal facial-motion clip.}
\label{tab:dataset_comparison}
\resizebox{\textwidth}{!}{%
\begin{tabular}{lcc>{\columncolor{lightgrayborder}}c}
\toprule
 & \textbf{Real ME$^\dagger$} & \textbf{MiE-X} & \textbf{Ours} \\
\midrule
\multicolumn{4}{l}{\textit{Scale \& Diversity}} \\ \hline
Number of human subjects & 16--32 per dataset & -- & \cellcolor{lightgrayborder}-- \\
Source images & -- & 5K & \cellcolor{lightgrayborder}15K \\
Total samples & 159--247 per dataset & 45K & \cellcolor{lightgrayborder}75K \\
Samples per class & Imbalanced & Imbalanced & \cellcolor{lightgrayborder}15K \\
\midrule
\multicolumn{4}{l}{\textit{Data Characteristics}} \\ \hline
Temporal (video) & \cmark & \xmark & \cellcolor{lightgrayborder}\cmark \\
Generated temporal facial-motion clip$^\ddagger$ & \cmark & \xmark & \cellcolor{lightgrayborder}\cmark \\
Resolution & 640$\times$480 & 128$\times$128 & \cellcolor{lightgrayborder}256$\times$256 \\
\midrule
\multicolumn{4}{l}{\textit{Emotion Classes}} \\ \hline
Number of classes & 3--7 & 3 & \cellcolor{lightgrayborder}5 \\
Categories & Varied$^*$ & Pos, Neg, Sur & \cellcolor{lightgrayborder}Hap, Sad, Sur, Dis, Ang \\
\midrule
\multicolumn{4}{l}{\textit{Controllability \& Annotations}} \\ \hline
Source-face selection & \xmark & \xmark & \cellcolor{lightgrayborder}\cmark \\
Estimated demographic labels & \xmark & \xmark & \cellcolor{lightgrayborder}\cmark \\
Estimated demographic coverage & Limited & N/A & \cellcolor{lightgrayborder}6 estimated demographic categories, 6 estimated age groups \\
Semantic metadata & \xmark & \xmark & \cellcolor{lightgrayborder}\cmark \\
\bottomrule
\end{tabular}}
\vspace{1mm}
\raggedright
\footnotesize{$^*$Hap, Sad, Sur, Dis, Ang, Fear, Contempt (varies by dataset). $^\ddagger$Multi-frame facial-motion sequence.}
\end{table*}

\section{Related Work}

\subsection{Micro-Expression Datasets}
MER research has long been constrained by the size and collection difficulty of real-world datasets. Widely used benchmarks such as CASME II, CAS(ME)$^2$, SAMM, and SMIC provide valuable spontaneous micro-expression recordings, but they remain small and are collected under constrained laboratory settings \cite{casmeii, casmesquare, casmecube, samm, sammlv, smic, smice}. As a result, models trained on these datasets often face large domain shifts and unstable cross-dataset performance.

\subsection{Synthetic Data for Micro-Expressions}
Several studies have explored synthetic augmentation for MER, including image-based micro-expression synthesis and AU-conditioned generation \cite{yap2021synthesising, zhang2023facial, zhao2022fine}. More recently, MiE-X introduced a large synthetic resource built from in-the-wild source images \cite{miex}. These efforts suggest that synthetic data can improve coverage and controllability, but available datasets still leave open problems in temporal modeling, video realism, and demographic analysis. EquiME extends this line of work by targeting full micro-expression video generation rather than frame pairs and by explicitly retaining demographic metadata for downstream evaluation.

\subsection{Action Units and Structured Conditioning}
The Facial Action Coding System (FACS) provides an interpretable representation of facial muscle activations and is widely used in affective computing \cite{facs}. For MER, AUs are especially useful because they encode subtle local movements more directly than coarse emotion labels alone. We therefore use AUs as controllable variables in the generation pipeline. This gives the synthesis process a clearer structure than prompt-only generation and is intended to align each video with its target emotional content.

\begin{figure*}[t]
\centering
\includegraphics[width=.7\textwidth]{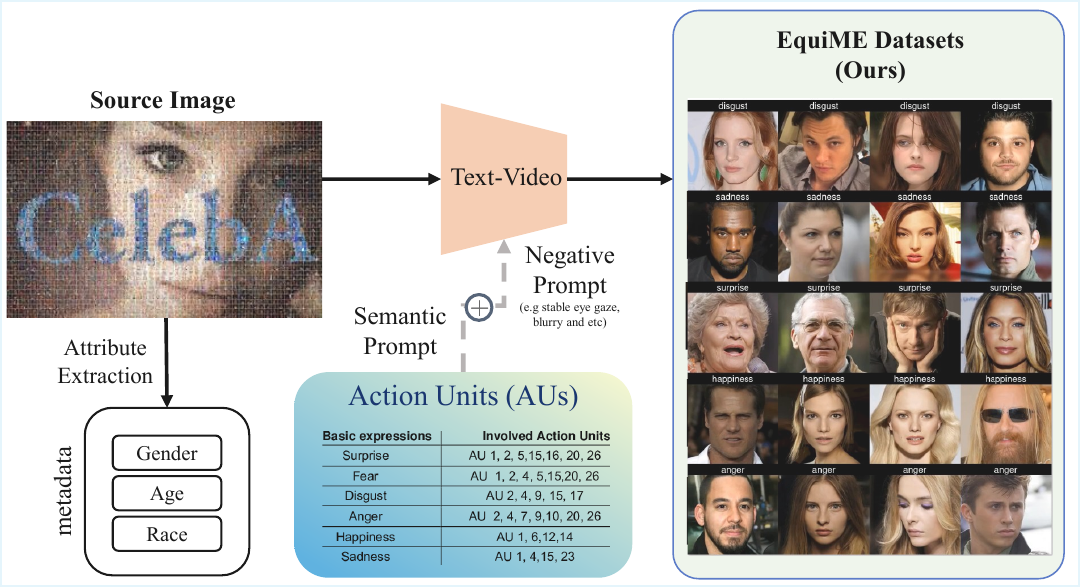}
\caption{Overview of the EquiME generation pipeline. A source face image, AU-conditioned text prompt, and negative prompt are passed to an image-to-video model to synthesize a short micro-expression clip.}
\label{fig:pipeline}
\end{figure*}

\section{EquiME Generation Protocol}

\subsection{Source Images and Design Goals}
We build EquiME from CelebA-HQ \cite{CelebAMask-HQ}, which provides high-quality face images with substantial variation in apparent identity, age appearance, and visual conditions. The goal is not to reproduce real-world MER collection exactly, but to create a scalable synthetic resource that preserves source-face diversity while enabling controlled generation of subtle facial motions. No additional preprocessing is applied before video synthesis.

\subsection{Demographic Label Extraction}
The public repository includes a DeepFace-based analysis script for estimating apparent demographic and facial attributes. In this pipeline, \texttt{DeepFace}\footnote{\url{https://github.com/serengil/deepface}} is applied to sampled video frames with the \texttt{age}, \texttt{gender}, \texttt{race}, and \texttt{emotion} actions enabled.

Source-face selection is guided by the CelebA-HQ attribute provided by the dataset, which provides \texttt{Male} and \texttt{Smiling} annotations for filtering source images. The DeepFace emotion output is not used as the dataset ground-truth emotion label; the released emotion classes are determined by the generation condition. We therefore treat these metadata as automatically inferred apparent attributes rather than self-identified personal attributes, and the predictions may contain errors.

\subsection{AU-Guided Image-to-Video Generation}
EquiME is generated with the LTX-Video image-to-video model \cite{HaCohen2024LTXVideo}. Starting from a single source image, we synthesize a short clip that depicts a target micro-expression-oriented facial motion while attempting to keep the subject identity stable over time. Figure~\ref{fig:pipeline} summarizes the workflow.

\subsubsection{Prompt Engineering for Emotion Control}
We use FACS-inspired prompts to specify the subtle muscle activations associated with each target emotion. Each prompt describes the relevant AU configuration and asks for a brief facial change rather than an exaggerated macro-expression. The released dataset covers five emotions: happiness, sadness, surprise, disgust, and anger. This AU-based prompting strategy is intended to improve controllability by grounding each class in visible muscle movements rather than relying on broad verbal labels alone.

\subsubsection{Negative Prompt Constraints}
We pair each positive prompt with a negative prompt that suppresses artifacts that would reduce MER usefulness, including motion blur, camera shake, distorted features, multiple faces, large head pose changes, open mouths, and exaggerated expressions \cite{ban2024understanding}. These constraints are intended to keep the synthesis focused on subtle facial motion and to reduce visually obvious failures.

\subsection{Structured AU-Conditioning Formulation}
In the implemented pipeline, the process starts from a target emotion label, maps that label to an AU-based facial movement description, and then generates a video conditioned on both the source image and the resulting prompt. This formulation represents the design structure of the generation pipeline and is not intended as a causal-identification analysis of human emotion.

\subsubsection{Formulation}
We define four variables: \(I\) for the source image, \(Y\) for the target emotion label, \(Z\) for the AU specification derived from that label, and \(X\) for the generated video. The prompt-construction stage is modeled as
\begin{equation}
Z = g(Y),
\end{equation}
where \(g(\cdot)\) denotes the explicit mapping from an emotion class to its AU-conditioned prompt description.

Video generation is then modeled as
\begin{equation}
X = G(I, Z, Y, \eta),
\end{equation}
where \(G(\cdot)\) is the image-to-video generator and \(\eta\) captures stochastic variation in the synthesis process. This formulation is intentionally simple, but it makes the implemented conditioning path explicit: a source face image is combined with a target emotion and its corresponding AU description to produce a prompted video clip.

\subsection{Dataset Structure}
The final release contains 75K videos generated from 15K source face images at \(256 \times 256\) resolution. Each raw clip contains 161 frames and is stored in MP4 format. In the current release, the MP4 files are stored at 24 fps, so each raw clip is approximately 6.7 seconds long. Because this raw duration is longer than a conventional spontaneous micro-expression, we refer to the generated samples as \emph{micro-expression-oriented synthetic facial-motion clips}. The generated videos are intended to contain a brief prompted facial-motion progression rather than a claim of physiologically verified spontaneous ME timing. For downstream MER training in our released baselines, the raw MP4 is converted to grayscale, resized to \(64 \times 64\), and truncated to the first 96 frames. We keep the first 96 frames because that is the input window used by the released baselines, not because the pipeline explicitly detects onset, apex, or recovery. Sequence models use this 96-frame tensor directly, while the MobileNetV2-based baseline uses only the middle frame of that tensor after conversion to three channels and resizing to \(224 \times 224\). Every sample is associated with an emotion label derived from the generation condition and automatically inferred demographic metadata used for aggregate analysis and subset construction.

\section{Evaluation}

\subsection{Experimental Setup}
We evaluate EquiME from two perspectives: video quality and downstream MER transfer. For downstream evaluation, we use SAMM \cite{samm}, CASME II \cite{casmeii}, and MiE-X \cite{miex} as comparison datasets.

SAMM is a high-resolution spontaneous micro-expression dataset with detailed FACS annotations. CASME II is a widely used benchmark with high frame-rate recordings collected in controlled conditions. MiE-X is a recent synthetic MER dataset that serves as the most relevant open synthetic baseline.

We report results for four baseline MER models with different capacity and temporal modeling behavior because we wanted to test both ME-specific architectures and lighter-weight classification models. ST-CNN \cite{microexpstcnn} is an ME-specific 3D spatiotemporal model that processes the full sequence. ResNet3D is implemented here as a 3D adaptation of ResNet \cite{resnet}. The compact 3DCNN is a small sequence model inspired by C3D \cite{3dcnn}. The MobileNetV2-based hybrid model uses a lightweight image backbone \cite{mobilenetv2} and operates on a reduced frame representation. All models are trained with categorical cross-entropy and Adam using a learning rate of \(10^{-4}\). For held-out EquiME experiments, we use a sample-level random 80/20 train/validation split with random seed 4, batch size 16, and a maximum of 100 epochs. The split is performed by video sample rather than by source face, so the same source face can appear in both partitions. Training uses model checkpointing on validation accuracy together with early stopping on validation loss (patience 15, restore best weights) and learning-rate reduction on validation loss (factor 0.2, patience 5, minimum learning rate \(10^{-5}\)). Table~\ref{tab:class_performance} reports this held-out EquiME split, while Table~\ref{tab:consistency_colored} reports cross-dataset transfer.

For label harmonization, the five-class setting uses happiness, sadness, surprise, disgust, and anger. The three-class setting maps labels to Positive / Negative / Surprise, where Positive = happiness, Negative = sadness + disgust + anger + fear when that label is present in an external dataset, and Surprise = surprise. For CASME II, the available ``repression'' label is grouped into Negative in the released cross-dataset code. Cross-dataset evaluations use this harmonized mapping across EquiME, SAMM, CASME II, and MiE-X.

\subsection{Video Quality}
Table~\ref{tab:baseline_results} compares EquiME with real and synthetic datasets using PSNR, SSIM, TV, BRISQUE, and CLIP-IQA \cite{kastryulin2022piq}. We frame these as frame-pair similarity and perceptual-quality metrics rather than as direct fidelity scores. PSNR and SSIM are computed on consecutive grayscale frame pairs within each video. TV denotes spatial total variation computed independently on each grayscale frame. BRISQUE and CLIP-IQA are used as no-reference perceptual-quality measures on individual frames. Values are averaged first per video and then across each dataset. For MiE-X, the available paired onset-apex images are treated as frame pairs for the same metric code path. EquiME does not dominate every metric, but within this comparison it shows a favorable trade-off across the selected no-reference perceptual-quality measures. In particular, EquiME obtains lower BRISQUE and higher CLIP-IQA than MiE-X under the selected no-reference perceptual-quality measures.
\begin{table*}[t]
\centering
\caption{Frame-pair similarity and perceptual-quality metrics on real and synthetic datasets.$^\dagger$}
\label{tab:baseline_results}
\resizebox{\textwidth}{!}{%
\begin{tabular}{lccccc}
\toprule
\textbf{Dataset} & \textbf{PSNR} & \textbf{SSIM} & \textbf{TV} & \textbf{BRISQUE}$\downarrow$ & \textbf{CLIP-IQA}$\uparrow$ \\
\midrule
\multicolumn{6}{c}{\textit{Real Datasets}} \\
\midrule
SAMM & 39.15 & 0.9167 & 26.40 & 7.96 & 0.48 \\
SMIC & 39.94 & 0.9460 & 15.22 & 34.47 & 0.25 \\
\rowcolor{gray!10}
Avg (Real) & 39.55 & 0.9314 & 20.81 & 21.22 & 0.365 \\
\midrule
\multicolumn{6}{c}{\textit{Synthetic Datasets}} \\
\midrule
MiE-X & 42.49 & 0.9787 & 6.96 & 27.28 & 0.61 \\
\rowcolor{gray!10}
Ours & \textbf{41.03} & \textbf{0.9684} & \textbf{13.80} & \textbf{14.66} & \textbf{0.79} \\
\midrule
\multicolumn{6}{c}{\textit{Difference from Avg (Real)}} \\
\midrule
MiE-X & +2.94 & +0.0473 & --13.85 & +6.06 & +0.245 \\
\rowcolor{gray!10}
Ours & \textbf{+1.48} & \textbf{+0.0370} & \textbf{--7.01} & \textbf{--6.56} & \textbf{+0.425} \\
\bottomrule
\end{tabular}%
}
\vspace{1mm}
\raggedright
\footnotesize{Higher PSNR and SSIM indicate greater similarity between consecutive frames; lower TV indicates lower within-frame spatial variation. These measures should not be interpreted as direct measures of expression realism. For BRISQUE, lower is better; for CLIP-IQA, higher is better. Best synthetic results are shown in \textbf{bold}.}
\end{table*}

\subsection{MER Performance}
Table~\ref{tab:class_performance} summarizes held-out EquiME performance for 3-class and 5-class tasks. As expected, the 5-class setting is harder. In the three-class setting, MobileNetV2 and the compact 3DCNN obtain the highest weighted F1 scores. Mean confidence denotes the average softmax confidence assigned to the predicted class and is reported as a descriptive model-behavior statistic rather than as a stand-alone quality metric. The available five-class evaluation includes ResNet3D and MobileNetV2; therefore, Table~\ref{tab:class_performance} reports five-class results only for those two architectures.

\begin{table*}[t]
\centering
\caption{Held-out EquiME performance on 3-class and 5-class tasks.}
\resizebox{\textwidth}{!}{%
\begin{tabular}{lcccc}
\hline
\textbf{Model} & \textbf{Classes} & \textbf{Weighted F1} & \textbf{Accuracy} & \textbf{Mean Confidence} \\
\hline
ST-CNN       & 3 & 0.49 & 0.63 & 0.60 \\
ResNet3D     & 3 & 0.61 & 0.48 & 0.60 \\
3DCNN        & 3 & 0.63 & 0.49 & 0.68 \\
MobileNetV2 & 3 & 0.63 & 0.54 & 0.63 \\
\hline
ResNet3D     & 5 & 0.36 & 0.42 & 0.56 \\
MobileNetV2 & 5 & 0.45 & 0.47 & 0.36 \\
\hline
\end{tabular}%
}
\label{tab:class_performance}
\end{table*}

The more important test for a synthetic dataset is transfer. Table~\ref{tab:consistency_colored} shows cross-dataset weighted F1-scores when models are trained on one dataset and tested on another. In this comparison, EquiME performs similarly to real-dataset training on several settings and varies less across architectures than the other training sets reported here. We define the architecture stability score as
\begin{equation}
\mathrm{Architecture\ Stability} = 1 - \mathrm{Std}\left(\mathrm{F1}_1,\mathrm{F1}_2,\mathrm{F1}_3,\mathrm{F1}_4\right),
\end{equation}
where the standard deviation is taken over the four architecture-specific weighted F1-scores in each train$\rightarrow$test row. Higher values are better, and values closer to 1 indicate lower variation across architectures. This score measures only variation across architectures and does not measure average predictive performance, so it should be interpreted together with the individual F1 scores. EquiME ties for the highest architecture stability when SAMM is the target and obtains the highest score when CASME II is the target.

\begin{table*}[t]
\centering
\caption{Cross-dataset F1-scores and architecture stability scores.$^\dagger$}
\label{tab:consistency_colored}
\resizebox{\textwidth}{!}{%
\begin{tabular}{lccccc}
\toprule
\textbf{Train $\rightarrow$ Test} & \textbf{ST-CNN} & \textbf{3DCNN} & \textbf{ResNet} & \textbf{MobileNetV2} & \textbf{Architecture Stability} \\
\midrule
\rowcolor{lightgrayborder} Ours$\rightarrow$SAM   & 0.489 & 0.488 & 0.481 & \textbf{0.537} & \textbf{0.978} \\
\rowcolor{lightgrayborder} Ours$\rightarrow$CAS   & 0.486 & \textbf{0.493} & 0.477 & 0.490 & \textbf{0.994} \\
\midrule
CAS$\rightarrow$SAM    & 0.526 & \textbf{0.562} & 0.518 & 0.501 & 0.978 \\
MiE-X$\rightarrow$SAM  & 0.490 & 0.085 & 0.482 & 0.466 & 0.854 \\
SAM$\rightarrow$CAS    & 0.454 & 0.326 & 0.474 & \textbf{0.494} & 0.938 \\
MiE-X$\rightarrow$CAS  & 0.525 & 0.524 & 0.405 & \textbf{0.526} & 0.951 \\
\bottomrule
\end{tabular}%
}
\vspace{1mm}
\raggedright
\footnotesize{$^\dagger$SAM = SAMM; CAS = CASME II. Best F1-score per row is shown in \textbf{bold}.}
\end{table*}

\subsection{Discussion}
The results support two main observations. First, EquiME is better framed as a transfer resource than as a claim of perfect realism. The quality metrics indicate that the videos are plausible and relatively clean, while the cross-dataset results indicate that models trained on EquiME can transfer to the real benchmarks used here. Second, the dataset appears to reduce variation across model families in these experiments, which is relevant for MER because current results are often brittle and architecture-dependent.

\section{Ethical Considerations and Limitations}
EquiME is intended as a research dataset, but micro-expression analysis raises non-trivial concerns around privacy, consent, and misuse. Systems built for emotional inference may be deployed in surveillance or decision-making contexts where subjects have limited awareness or no meaningful ability to opt out. We therefore frame EquiME as a resource for studying MER generalization and dataset effects, not as evidence that emotion can be inferred reliably in high-stakes settings.

The dataset has several technical limitations. Although EquiME expands demographic coverage relative to common MER benchmarks, its distribution is still uneven, especially across the automatically inferred demographic and age groups. In addition, synthetic generation can import biases from both the source-image collection and the underlying video model. AU-based prompting improves controllability, but it does not guarantee that all generated expressions are equally realistic across the automatically inferred demographic groups or cultures.

Privacy risks also matter at deployment time. If MER systems are used in real environments, they may enable unwanted inference of subtle affective signals. One possible mitigation direction is privacy-preserving intervention at the input level, including adversarial perturbations or physical countermeasures. Our prior work explored post-hoc adversarial stickers against micro-expression leakage as one such defense strategy \cite{10889387}. We do not evaluate these defenses in the present paper, but they are relevant to the broader responsible-use context for MER systems.

To reduce downstream misuse, the dataset is released under a Creative Commons Attribution-NonCommercial-ShareAlike 4.0 International License (CC BY-NC-SA 4.0) to communicate intended non-commercial research use, although licensing alone cannot prevent misuse. We also recommend that future work evaluate demographic slices explicitly, document failure cases, and avoid overstating the interpretability of micro-expression-based emotion inference.

\section{Conclusion}
EquiME addresses a practical bottleneck in micro-expression research: the lack of scalable and temporally coherent training data. We introduced a synthetic video dataset built with AU-guided image-to-video generation, together with a structured AU-conditioning formulation that makes the control strategy explicit. Across perceptual quality analysis and cross-dataset MER experiments, the results suggest that synthetic videos can serve as a useful training resource for transfer to the real benchmarks considered here.

Our central claim is deliberately narrow: EquiME is intended as a scalable synthetic resource for studying cross-dataset transfer in micro-expression recognition. It complements rather than replaces spontaneous, consented real-world datasets. Its automatically inferred demographic labels, synthetic expression dynamics, and uneven group representation should be considered when interpreting downstream results.

{
    \small
    \bibliographystyle{ieeenat_fullname}
    \bibliography{ref}

@inproceedings{10889387,
  author={Tan, Pei-Sze and Rajanala, Sailaja and Tan, Yee-Fan and Pal, Arghya and Tan, Chun-Ling and Phan, Raphaël C.-W. and Ong, Huey-Fang},
  booktitle={ICASSP 2025 - 2025 IEEE International Conference on Acoustics, Speech and Signal Processing (ICASSP)}, 
  title={Post-Hoc Adversarial Stickers Against Micro-Expression Leakage}, 
  year={2025},
  pages={1-5},
  doi={10.1109/ICASSP49660.2025.10889387}}

@article{dominguez2024metrics,
  title={Metrics for dataset demographic bias: A case study on facial expression recognition},
  author={Dominguez-Catena, Iris and Paternain, Daniel and Galar, Mikel},
  journal={IEEE Transactions on Pattern Analysis and Machine Intelligence},
  volume={46},
  number={8},
  pages={5209--5226},
  year={2024},
  publisher={IEEE}
}

@inproceedings{mobilenetv2,
  title={MobileNetV2: Inverted Residuals and Linear Bottlenecks},
  author={Sandler, Mark and Howard, Andrew and Zhu, Menglong and Zhmoginov, Andrey and Chen, Liang-Chieh},
  booktitle={Proceedings of the IEEE/CVF Conference on Computer Vision and Pattern Recognition (CVPR)},
  pages={4510--4520},
  year={2018}
}

@inproceedings{3dcnn,
  title={Learning spatiotemporal features with 3D convolutional networks},
  author={Tran, Du and Bourdev, Lubomir and Fergus, Rob and Torresani, Lorenzo and Paluri, Manohar},
  booktitle={Proceedings of the IEEE international conference on computer vision (ICCV)},
  year={2015}
}

@inproceedings{ban2024understanding,
  title={Understanding the Impact of Negative Prompts: When and How Do They Take Effect?},
  author={Ban, Yuanhao and Wang, Ruochen and Zhou, Tianyi and Cheng, Minhao and Gong, Boqing and Hsieh, Cho-Jui},
  booktitle={European Conference on Computer Vision},
  pages={190--206},
  year={2024},
  organization={Springer}
}

@inproceedings{xu2020investigating,
  title={Investigating bias and fairness in facial expression recognition},
  author={Xu, Tian and White, Jennifer and Kalkan, Sinan and Gunes, Hatice},
  booktitle={Computer Vision--ECCV 2020 Workshops: Glasgow, UK, August 23--28, 2020, Proceedings, Part VI 16},
  pages={506--523},
  year={2020},
  organization={Springer}
}

@misc{kastryulin2022piq,
  title = {PyTorch Image Quality: Metrics for Image Quality Assessment},
  url = {https://arxiv.org/abs/2208.14818},
  author = {Kastryulin, Sergey and Zakirov, Jamil and Prokopenko, Denis and Dylov, Dmitry V.},
  doi = {10.48550/ARXIV.2208.14818},
  publisher = {arXiv},
  year = {2022}
}

@article{zhang2023facial,
  title={Facial prior guided micro-expression generation},
  author={Zhang, Yi and Xu, Xinhua and Zhao, Youjun and Wen, Yuhang and Tang, Zixuan and Liu, Mengyuan},
  journal={IEEE Transactions on Image Processing},
  volume={33},
  pages={525--540},
  year={2023},
  publisher={IEEE}}

@inproceedings{zhao2022fine,
  title={Fine-grained micro-expression generation based on thin-plate spline and relative au constraint},
  author={Zhao, Sirui and Yin, Shukang and Tang, Huaying and Jin, Rijin and Xu, Yifan and Xu, Tong and Chen, Enhong},
  booktitle={Proceedings of the 30th ACM International Conference on Multimedia},
  pages={7150--7154},
  year={2022}
}

@article{xie2022overview,
  title={An overview of facial micro-expression analysis: Data, methodology and challenge},
  author={Xie, Hong-Xia and Lo, Ling and Shuai, Hong-Han and Cheng, Wen-Huang},
  journal={IEEE Transactions on Affective Computing},
  volume={14},
  number={3},
  pages={1857--1875},
  year={2022},
  publisher={IEEE}
}

@article{yap2021synthesising,
  title={Synthesising facial macro-and micro-expressions using reference guided style transfer},
  author={Yap, Chuin Hong and Cunningham, Ryan and Davison, Adrian K and Yap, Moi Hoon},
  journal={Journal of Imaging},
  volume={7},
  number={8},
  pages={142},
  year={2021},
  publisher={MDPI}
}

@article{HaCohen2024LTXVideo,
  title={LTX-Video: Realtime Video Latent Diffusion},
  author={HaCohen, Yoav and Chiprut, Nisan and Brazowski, Benny and Shalem, Daniel and Moshe, Dudu and Richardson, Eitan and Levin, Eran and Shiran, Guy and Zabari, Nir and Gordon, Ori and Panet, Poriya and Weissbuch, Sapir and Kulikov, Victor and Bitterman, Yaki and Melumian, Zeev and Bibi, Ofir},
  journal={arXiv preprint arXiv:2501.00103},
  year={2025}
}

@misc{microexpstcnn,
  doi = {10.48550/ARXIV.1904.01390},
  
  url = {https://arxiv.org/abs/1904.01390},
  
  author = {Reddy, Sai Prasanna Teja and Karri, Surya Teja and Dubey, Shiv Ram and Mukherjee, Snehasis},
  
  title = {Spontaneous Facial Micro-Expression Recognition using 3D Spatiotemporal Convolutional Neural Networks},
  
  publisher = {arXiv},
  
  year = {2019},
  
  copyright = {arXiv.org perpetual, non-exclusive license}
}

@article{austin2025distributional,
  title={Distributional bias compromises leave-one-out cross-validation},
  author={Austin, George I and Pe'er, Itsik and Korem, Tal},
  journal={arXiv preprint arXiv:2406.01652},
  year={2024}
}

@article{gronau2019limitations,
  title={Limitations of Bayesian leave-one-out cross-validation for model selection},
  author={Gronau, Quentin F and Wagenmakers, Eric-Jan},
  journal={Computational brain \& behavior},
  volume={2},
  number={1},
  pages={1--11},
  year={2019},
  publisher={Springer}
}

@inproceedings{miex,
  title={How to Synthesize a Large-Scale and Trainable Micro-Expression Dataset?},
  author={Liu, Yuchi and Wang, Zhongdao and Gedeon, Tom and Zheng, Liang},
  booktitle={ECCV},
  year={2022}
}

@ARTICLE{smice,
  title={Micro-expression spotting: A new benchmark},
  author={Tran, Thuong-Khanh and Vo, Quang-Nhat and Hong, Xiaopeng and Li, Xiaobai and Zhao, Guoying},
  journal={Neurocomputing},
  volume={443},
  pages={356--368},
  year={2021},
  publisher={Elsevier}
}

@ARTICLE{samm,
  title={Samm: A spontaneous micro-facial movement dataset},
  author={Davison, Adrian K and Lansley, Cliff and Costen, Nicholas and Tan, Kevin and Yap, Moi Hoon},
  journal={IEEE transactions on affective computing},
  volume={9},
  number={1},
  pages={116--129},
  year={2016},
  publisher={IEEE}
}

@ARTICLE{casmecube,
  title={CAS (ME) 3: A third generation facial spontaneous micro-expression database with depth information and high ecological validity},
  author={Li, Jingting and Dong, Zizhao and Lu, Shaoyuan and Wang, Su-Jing and Yan, Wen-Jing and Ma, Yinhuan and Liu, Ye and Huang, Changbing and Fu, Xiaolan},
  journal={IEEE Transactions on Pattern Analysis and Machine Intelligence},
  year={2022},
  publisher={IEEE}
}

@ARTICLE{casmesquare,
  title={{CAS(ME)$^2$}: A Database for Spontaneous Macro-Expression and Micro-Expression Spotting and Recognition},
  author={Qu, Fangbing and Wang, Su-Jing and Yan, Wen-Jing and Li, He and Wu, Shuhang and Fu, Xiaolan},
  journal={IEEE Transactions on Affective Computing},
  volume={9},
  number={4},
  pages={424--436},
  year={2017},
  publisher={IEEE}
}

@inproceedings{smic,
  title={A spontaneous micro-expression database: Inducement, collection and baseline},
  author={Li, Xiaobai and Pfister, Tomas and Huang, Xiaohua and Zhao, Guoying and Pietik{\"a}inen, Matti},
  booktitle={2013 10th IEEE International Conference and Workshops on Automatic face and gesture recognition (fg)},
  pages={1--6},
  year={2013},
  organization={IEEE}
}

@ARTICLE{casmeii,
    author = {Yan, Wen-Jing AND Li, Xiaobai AND Wang, Su-Jing AND Zhao, Guoying AND Liu, Yong-Jin AND Chen, Yu-Hsin AND Fu, Xiaolan},
    publisher = {Public Library of Science},
    journal = {PLOS ONE},
    title = {CASME II: An Improved Spontaneous Micro-Expression Database and the Baseline Evaluation},
    year = {2014},
    month = {01},
    volume = {9},
    url = {https://doi.org/10.1371/journal.pone.0086041},
    pages = {1-8},
}

@ARTICLE{surveydeep,
  author={Li, Yante and Wei, Jinsheng and Liu, Yang and Kauttonen, Janne and Zhao, Guoying},
  journal={IEEE Transactions on Affective Computing}, 
  title={Deep Learning for Micro-Expression Recognition: A Survey}, 
  year={2022},
  volume={13},
  number={4},
  pages={2028-2046},
  doi={10.1109/TAFFC.2022.3205170}}

@inproceedings{sammlv,
author = {Yap, Chuin Hong and Kendrick, Connah and Yap, Moi Hoon},
year = {2020},
month = {11},
booktitle = {2020 15th IEEE International Conference on Automatic Face and Gesture Recognition (FG 2020)},
pages = {771-776},
title = {SAMM Long Videos: A Spontaneous Facial Micro- and Macro-Expressions Dataset},
doi = {10.1109/FG47880.2020.00029}
}

@inproceedings{CelebAMask-HQ,
  title={MaskGAN: Towards Diverse and Interactive Facial Image Manipulation},
  author={Lee, Cheng-Han and Liu, Ziwei and Wu, Lingyun and Luo, Ping},
  booktitle={IEEE Conference on Computer Vision and Pattern Recognition (CVPR)},
  year={2020}
}

@inproceedings{resnet,
  title={Deep residual learning for image recognition},
  author={He, Kaiming and Zhang, Xiangyu and Ren, Shaoqing and Sun, Jian},
  booktitle={Proceedings of the IEEE conference on computer vision and pattern recognition},
  pages={770--778},
  year={2016}
}

@book{facs,
    author = {Ekman, Paul and Rosenberg, Erika L.},
    title = "{What the Face Reveals: Basic and Applied Studies of Spontaneous Expression Using the Facial Action Coding System (FACS)}",
    publisher = {Oxford University Press},
    year = {2005},
    month = {04},
    isbn = {9780195179644},
    doi = {10.1093/acprof:oso/9780195179644.001.0001},
    url = {https://doi.org/10.1093/acprof:oso/9780195179644.001.0001},
}
}

\end{document}